\documentclass{article}

\usepackage{arxiv}

\usepackage[utf8]{inputenc} 
\usepackage[T1]{fontenc}    
\usepackage{hyperref}       
\usepackage{url}            
\usepackage{booktabs}       
\usepackage{amsfonts}       
\usepackage{nicefrac}       
\usepackage{microtype}      
\usepackage{lipsum}		
\usepackage{graphicx}
\usepackage{natbib}
\usepackage{doi}

\usepackage{mathtools}
\usepackage{amsthm}
\usepackage[noend]{algorithmic}
\usepackage{algorithm}
\theoremstyle{plain}

\theoremstyle{definition}

\theoremstyle{remark}

\usepackage{multirow}
\usepackage[table]{xcolor}
\usepackage{color, colortbl, xcolor}
\definecolor{Gray}{gray}{0.9}
\definecolor{light-gray}{gray}{0.95}

\usepackage{listings}
\usepackage{fvextra}
\usepackage{array}
\usepackage[labelsep=period]{caption}
\newcommand{\ours}{HouseTune}
\usepackage{tabularx}
\usepackage{booktabs}
\RequirePackage{subcaption}

\title{HouseTune: Two-Stage Floorplan Generation with LLM Assistance}

\author{Ziyang Zong\\
Sun Yat-sen University\\
{\tt\small zongzy@mail2.sysu.edu.cn}
\and
Guanying Chen\\
Sun Yat-sen University\\
{\tt\small guanying2018@gmail.com}
\and
Zhaohuan Zhan\\
Sun Yat-sen University\\
{\tt\small zhanzhh5@mail2.sysu.edu.cn}
\and
Fengcheng Yu\\
Sun Yat-sen University\\
{\tt\small yufch3@mail2.sysu.edu.cn}
\and
Guang Tan\\
Sun Yat-sen University\\
{\tt\small tanguang@mail.sysu.edu.cn}
}

\renewcommand{\shorttitle}{\textit{arXiv} Template}

\begin{document}
\maketitle

\begin{abstract}
    This paper proposes a two-stage text-to-floorplan generation framework that combines the reasoning capability of Large Language Models (LLMs) with the generative power of diffusion models. In the first stage, we leverage a Chain-of-Thought (CoT) prompting strategy to guide an LLM in generating an initial layout (Layout-Init) from natural language descriptions, which ensures a user-friendly and intuitive design process. However, Layout-Init may lack precise geometric alignment and fine-grained structural details. To address this, the second stage employs a conditional diffusion model to refine Layout-Init into a final floorplan (Layout-Final) that better adheres to physical constraints and user requirements. Unlike prior methods, our approach effectively reduces the difficulty of floorplan generation learning without the need for extensive domain-specific training data. Experimental results demonstrate that our approach achieves state-of-the-art performance across all metrics, which validates its effectiveness in practical home design applications. 
Our code will be made publicly available. 
\end{abstract}

In architectural design, creating floorplans that align with user requirements remains a challenge.
Traditional design methods not only rely on specialized knowledge but also require designers to make iterative adjustments to accommodate specific user needs, which makes personalized design challenging.
Learning-based models~\cite{murali2017indoor,merrell2010computer,sun2022wallplan,luo2022floorplangan} have made efforts to improve the accuracy, interactivity, and efficiency of floorplan generation. Despite these advances, existing approaches have yet to achieve a level of user-friendliness and accuracy that enables ready adoption by ordinary users. 
 
Existing solutions~\cite{nauata2020house, nauata2021house, shabani2023housediffusion} typically treat the floorplan creation task as a conditional generation problem. The conditions are expressed through a bubble diagram, where rooms are represented as nodes (or ``bubbles''), and doors as edges specifying the spatial relationships between rooms, as shown in Figure~\ref{fig:introduction}(a). While the bubble diagram provides a precise way of describing the house layout, specifying a consistent graph structure can still be too demanding for novice users who wish to explore designing with minimal effort. We believe that a more user-friendly interface should allow users to express their needs in natural language. For example, the user could simply specify {\em ``I need a house with three bedrooms, a living room, a bathroom, a kitchen, and a balcony adjacent to the living room.''}
Alternatively, the user may use a simple menu to make choices, which are then translated into natural language. The text-to-floorplan paradigm  make the design process more accessible and intuitive for non-expert users.

A solution to this requirement involves training a generative model capable of directly mapping textual descriptions to house layouts, as depicted in Figure~\ref{fig:introduction}(b). The Tell2Design method~\cite{leng2023Tell2Design} addresses this task using a Sequence-to-Sequence approach, where the input text specifies the floorplan boundary and the exact geometry of individual rooms. The method imposes stringent demands for geometric detail and consistency, placing an even greater burden on users compared to working with bubble diagrams. In addition, the design limits the diversity of the results, which is often crucial for exploratory design.

\section{Introduction}
\label{sec:introduction}
\begin{figure*}
  \centering
     \includegraphics[width=\linewidth]{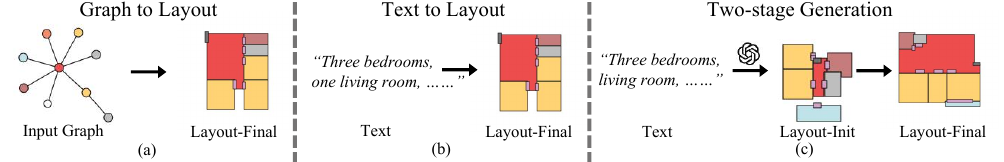}
     \caption{Comparison of different floorplan generation pipelines. (a) Graph-to-floorplan approach (e.g., HouseDiffusion), where rooms are represented as nodes and doors as edges, forming a graph~\cite{zhuo2022efficient, zhuo2022proximity, zhuo2024partitioning} that represents spatial relationships; (b) Text-to-floorplan approach, which directly maps a natural language description to a house layout; (c) Our two-stage pipeline, where an LLM is used to generate an initial layout, Layout-Init, according to the user's textual specification. The initial solution serves as a condition for generating the final layout, Layou-Final, through a diffusion model.}
     \label{fig:introduction}
\end{figure*}

We introduce \ours, a novel two-stage floorplan generation framework. In the first stage, we use a multimodal large language model (LLM) to generate an initial house layout, termed {\bf Layout-Init}.
The LLM’s common-sense knowledge and reasoning capability enable the creation of a preliminary floorplan based on user input.
In the second stage, a diffusion model refines this initial design into a more precise and reasonable final layout, referred to as {\bf Layout-Final}. This two-stage generation process is illustrated in Figure~\ref{fig:introduction}(c).

Generating a reasonable house layout using an LLM is not straightforward, as LLMs often struggle to interpret exact numeric and geometric constraints. To address this, we employ a Chain-of-Thought (CoT) prompting strategy~\cite{wei2022chain, zhang2022automatic}. This approach ensures the generated layout aligns with the user’s core requirements, such as the number of rooms, room types, and approximate spatial arrangements.

The initial layouts produced at this stage typically exhibit imperfections in object sizing and alignment, due to the inherent limitations of LLMs in handling intricate geometric details. To overcome this, we design a diffusion model that refines the initial layout into a final floorplan. While inspired by HouseDiffusion~\cite{shabani2023housediffusion}, our approach differs in its conditioning strategy: whereas HouseDiffusion applies conditioning solely during the reverse process, our model integrates conditioning in both the forward (noise addition) and reverse processes. This dual-conditioning mechanism ensures enhanced alignment with the specified constraints, yielding more precise and robust results.

We validate our approach on the RPlan dataset, in comparison with state-of-the-art methods.  For instance, in comparison with HouseDiffusion, our method achieves the best performance across all metrics, and in particular, with a 28\% improvement in diversity.
To summarize, this paper makes the following contributions:

\begin{itemize}
    \item We propose a novel two-stage floorplan generation framework that leverages the initial layouts generated by LLMs as the condition to generate the final floorplans.
    \item We develop a prompt design based on the Chain-of-Thought technique, successfully guiding the LLM to produce structured and coherent initial house layouts.
   \item We design a conditional diffusion model that refines the LLM-generated layouts into high-fidelity floorplans. Experimental results demonstrate that our method achieves state-of-the-art performance.
\end{itemize}
\section{Related Work}
\begin{figure*}
  \centering
    \includegraphics[width=\linewidth]{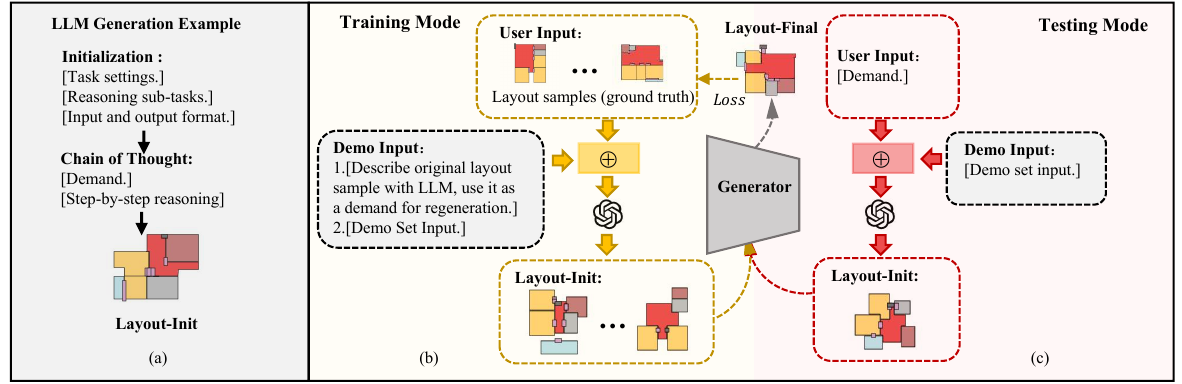}
    \caption{Training and testing processes of our method. (a) An example of LLM generating a Layout-Init according to user demands. (b) Given a house layout sample, we use the LLM to describe it. The description is used to mimic the user's demands. Using multiple examples as in (a) as demos, we ask the LLM to generate a Layout-Init for each sample. These initial layouts serve as conditions for the generator, which outputs Layouts-Final. (c) Given a textual description from the user, we again use the demos like (a) to obtain a Layout-Init, which goes through the diffusion model to generate Layout-Final.}
    \label{fig:LLM-intro}
  \hfill
\end{figure*}

\label{sec:reltedwork}
\textbf{Floorplan Generation.}
In the field of building design, generating high-quality house layouts has been an important research direction~\cite{hendrikx2013procedural, wu2018building, hu2020graph2plan,muller2006procedural,peng2014computing,sun2022wallplan}.
Nauata et al.~\cite{nauata2020house} proposed House-GAN, a method based on Generative Adversarial Networks (GANs) that achieves end-to-end automated house layout generation. Nauata et al.~\cite{nauata2021house} further proposed House-GAN++, which improved the original GAN structure by addressing the problems of missing doors.
Upadhyay et al.~\cite{upadhyay2022flnet} expanded on this by considering user inputs in the form of boundaries, room types, and spatial relationships. 
Hu et al.~\cite{hu2020graph2plan} presented Graph2Plan, which retrieves a set of floorplans with their associated layout graphs from a database, allowing a user to specify room counts and other layout constraints.
Shabani et al.~\cite{shabani2023housediffusion} introduced bubble diagrams as constraints and used a diffusion model to generate house layouts.
Chen et al.~\cite{chen2024polydiffuse} transformed visual sensor data into polygonal shapes with Diffusion Models.
Su et al.~\cite{su2024floor} developed a bi-directional structure of ``corruption and denoise'' approach to learn topology graphs.
Zeng et al.~\cite{zeng2024residential} proposed a multi-conditional two-stage generation model, which allows human designers to intervene and enhance controllability based on the denoising diffusion model.
Leng et al.~\cite{leng2023Tell2Design} introduced the Tell2Design (T2D) dataset, which comprises over 80,000 floorplan designs paired with natural language descriptions, in support of floorplan generation. HOLODECK~\cite{yang2024holodeck} explored the use of LLMs to generate floorplans populated with various objects, emphasizing the overall quality and consistency of the environments rather than dealing with complex floorplans.

\noindent\textbf{Conditional Diffusion.}
Conditional diffusion models are a subset of diffusion models~\cite{cao2024survey, sohl2015deep, yang2023diffusion} where the generation is conditioned on specific input data, such as labels, images, or text, allowing more control over the output~\cite{zhang2023adding, chen2022analog, ho2022video}.
Expanding on the success of diffusion models~\cite{sohl2015deep}, Ho et al. proposed DDPMs~\cite{song2020denoising}, introducing constraints within the diffusion process to guide generation, which led to significant significant improvements.
Subsequently, conditional diffusion models, such as DALLE-2~\cite{ramesh2022hierarchical} and Stable Diffusion~\cite{rombach2022high} have been applied to text-to-image generation.
Recent advancements continue to refine diffusion models’ conditional control. For instance,
Yang et al.~\cite{yang2024improving} improved diffuse image synthesis based on context prediction.
Yang et al.~\cite{yang2024lossy} proposed an end-to-end lossy image compression framework using conditional diffusion models to refine the source image.
The conditional diffusion model is a foundation of our approach, in which use it to refine the floorplan generated by the LLM.
\section{Method}

\ours\ uses a two-stage approach to generate house layouts. By breaking the generation pipeline into two stages, \ours\ manages to exploit the power of LLMs in interpreting user demands and generating approximate layouts with their common knowledge and reasoning ability. Further refinement is designed to enhance the layout's quality. Figure~\ref{fig:LLM-intro} depicts the training and testing processes of our method, in which Layout-Init plays a pivotal role.

\subsection{Layout-Init Generation}
\label{sec:initialsolutiofromllm}

\begin{figure*}
  \centering
    \includegraphics[width=\linewidth]{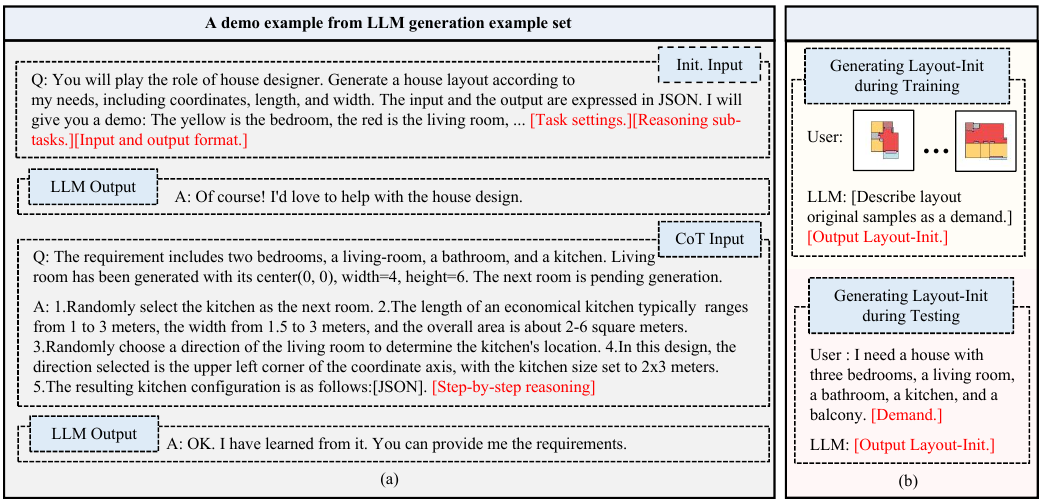}
    \caption{CoT-based prompting facilitates generation of Layout-Init and its function in training and testing. (a) An example showing how to interact with the LLM to obtain Layout-Init. The Initialization section defines the LLM's role as a house designer, and standardizes output format; the Chain of Thought section directs the LLM to create a house layout step by step, ensuring plausible room placement and sizing. (b) Invoking the Layout-Init generation during training and testing. }
    \label{fig:LLM-prompt}
  \hfill
\end{figure*}

Natural language descriptions alone may lack the detailed context necessary for accurate reasoning about specific house layouts~\cite{mann2020language, workrethinking}.
To address this limitation, we enhance the LLM's reasoning capability by allowing it to reference multiple human-designed layout demos, following the Chain of Thought (CoT) approach.
By demonstrating each step explicitly, including room selection, placement, and sizing decisions, CoT helps the LLM break down complex layout generation tasks into manageable, step-by-step inferences.

The demo set~$E$ for the prompt engineering consists of pairs of language description~$e_d$ and corresponding Layout-Init~$e_i$:
\begin{small}
\begin{equation}
    E=\{(e_i,e_d)_1,(e_i,e_d)_2,...,(e_i,e_d)_n\}.
\end{equation}
\end{small}
This demo set essentially encompasses diverse examples with varying house layout designs.
Figure~\ref{fig:LLM-intro}(a) shows the reasoning process of an LLM generation example.

Figures~\ref{fig:LLM-intro} (b), (c) illustrate the training and testing pipelines of our approach. In the testing mode, we input user prompts expressed in natural language and the demo set $E$ into the LLM to generate the initial layout. The output of the LLM is structured in JSON format, which can be described as follows:
\begin{small}
\begin{equation}
        \text{House}=\{\text{rooms}=[\text{Room}_\text{1},..., \text{Room}_\text{n}]\},
        \label{Formula:json_tree}    
        \vspace{-5pt}
\end{equation}
\end{small}
\begin{small}
\begin{equation}
    \begin{aligned}
        \text{Room}_\text{i} = \{ & \text{name}=[\text{Name}_\text{i}],\\
                    & \text{style}=\text{``modern''}, \\
                    & \text{position}=[(\text{X}_\text{i}, \text{Y}_\text{i})],\\ 
                    & \text{size}=[(\text{Width}_\text{i}, \text{Height}_\text{i})], \\
                    & \text{door}=[\text{Direction}_\text{i}] \},
    \end{aligned}
    \label{subsec:llm-format}
\end{equation}
\end{small}
where $\text{Direction}_i \in \{\text{``up''},\text{``down''},\text{``left''},\text{``right''}\}$.

The training process requires establishing a mapping from ground-truth layouts to Layout-Init, which are further used as conditions for generating Layout-Final. Given a house layout sample from the dataset, we first use the LLM to give a description of the layout, which mimics the user's demands. The description is then fed to the LLM, which generates Layout-Init using the CoT-based prompt.

Figure~\ref{fig:LLM-prompt} shows a sample of prompt design. The prompt used in our method consists of three parts: Initialization, Chain of Thought (CoT), and Layout-Init generation.

\noindent\textbf{Initialization.}
Initialization introduces task settings, input and output format specifications, and reasoning sub-tasks.
Specifically, the task utilizes the role-prompting technique~\cite{john2023art} to instruct the LLM to play the role of a designer: we present specific tasks to the LLM, prompting it to explain, step by step, the reasoning behind the design of each house location. An example prompt illustrating this setup is provided in see Figure~\ref{fig:LLM-prompt} (a).

\noindent\textbf{Chain of thought.} In each separate reasoning step, the LLM needs to determine the current house layout, then randomly select the next room to be generated, and infer the information for the next room based on the information from the existing rooms. The emphasis is on explaining the rationale behind these decisions.
Figure~\ref{fig:LLM-prompt} (a) presents an example that shows how to guide the model through each step of the inference process, including room selection, sizing, and positioning within the layout.

\noindent\textbf{Layout-Init generation.}
In this step, demos are input into the LLM to prompt the generation of the house layout. 
The user provides natural language input, instructs the LLM to perform a step-by-step reasoning process. The final output is structured in the JSON format.

\subsection{Diffusion Refinement for Layout-Final}

\label{sec:\ours}
\begin{figure*}[t]
  \centering
    \includegraphics[width=\linewidth]{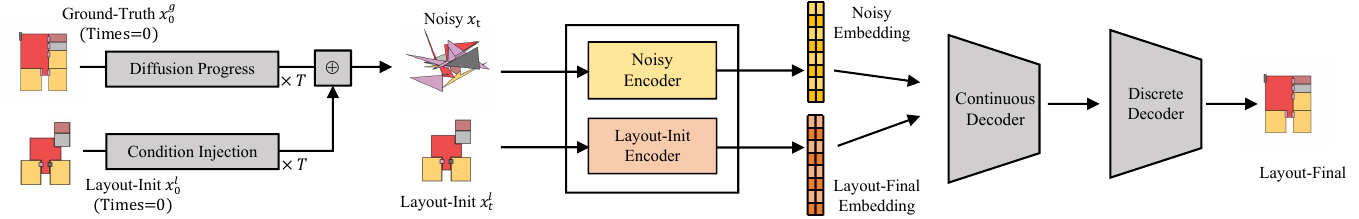}
    \caption{Conditional diffusion network for refining Layout-Init. The forward process takes the ground-truth house layout $x^g_0$ and the Layout-Init $x^i_0$ and adds a Gaussian noise to create a noisy house layout sample $x_T$. The reverse process takes a noisy house layout at time $t$ with Layout-Init as the condition. Two encoders are used to encode and obtain the latent representations for $x^g_t$ and $x^i_t$. }
    \label{fig:network}
  \hfill
\end{figure*}

Following HouseDiffusion~\cite{shabani2023housediffusion}, our conditional generation method is inspired by the diffusion model.
The reverse process consists of a set of modules: Layout-Init Encoder, Noisy Encoder, Continuous Decoder and Discrete Decoder.
Figure~\ref{fig:network} presents the specific training method of our model.

Unlike HouseDiffusion~\cite{shabani2023housediffusion}, which applies conditional information only during noise decoding, our approach fuses conditional information in both noise addition and decoding phases. By embedding Layout-Init from the start, the model builds condition-relevant features progressively during noise accumulation and continues to refine them in the denoising phase. This dual-phase conditioning enhances condition alignment and improves the stability and accuracy of the generated layouts.

\subsubsection{Diffusion Model}
\label{sec:Diffusion Model}
Diffusion models progressively refine Gaussian noise \begin{small}$x_T$\end{small} over \begin{small}$T$\end{small}, guided by a conditional Layout-Init \begin{small}$y$\end{small}. The forward process takes an initial data sample \begin{small}$x_0$\end{small}, a conditional Layout-Init \begin{small}$y$\end{small} and generates a noisy sample \begin{small}$x_t$\end{small} at time step \begin{small}$t$\end{small} by sampling Gaussian noise \begin{small}$\epsilon \sim N(0,1)$\end{small}, incorporating information from the conditional layout through its embedding \begin{small}$e(y)$\end{small}:
\begin{small}
   \begin{equation}
    x_t = \sqrt{\alpha_t}*x_0+\sqrt{1-\alpha_t}*\epsilon+e(y).
\end{equation}
\end{small}
$\alpha_t$ represents the noise schedule, controlling noise intensity changes from $1$ to $0$ at each time step $t$.
$e(y)$ is the embedding derived from the Layout-Init $y$, guiding the diffusion model to align with the features of the Layout-Init $y$.

The reverse denoising process starts with a fully noise-added sample $x_T$ and gradually removes noise to produce a sample that matches the target data distribution $x_0$.
This reverse process is typically implemented as an iterative process, where each time step depends on the output of the previous step and incorporates conditional information $y$ to guide the generation.
In the reverse denoising process at time step $t$, the model $f_{\theta}$ predicts the denoised image $\hat{x}_0$:
\begin{small}
   \begin{equation}
    \hat{x}_0 = f_{\theta}(x_t,t,e(y)).
\end{equation}
\end{small}
$f_{\theta}$ is the denoising network Transformer~\cite{vaswani2017attention} parameterized by $\theta$.
$x_t$ is the noisy image at current time step $t$.

The reverse process updates the state of $x_t$ each step to remove noise and gradually guide the generated image toward the structure of the condition $y$:
\begin{small}
   \begin{equation}
     x_{t-1} = \sqrt{\alpha_{t-1}}*\hat{x}_0+\sqrt{1-\alpha_{t-1}}*\epsilon.
\end{equation}
\end{small}
$\alpha_{t-1}$ is the noise level parameter at time step $t-1$. $\epsilon$ is the residual noise predicted by the model.

By iterating the update formula from time step $T$ to $1$, the reverse process progressively removes noise, bringing $x_t$ closer to a clean sample $x_0$.
At each time step, the embedding $e(y)$ from the condition $y$ continues to guide the generation, ensuring that the final output conforms to the structure and characteristics of the conditional layout.

\subsubsection{Network Architecture}
\label{sec:Network}
\textbf{Data representation.} Here we refer to the processing of data representation in~\cite{shabani2023housediffusion} to meet the constraints on the number and type of rooms.
Suppose \begin{small}$P^G$\end{small} is the set of polygonal loops for each room/door in the ground truth.
\begin{small}$P^{I}$\end{small} is for Layout-Init, and \begin{small}$P^F$\end{small} is for the LLM-Final. For \begin{small}$\forall P \in \{P^G,P^{I},P^F\}$\end{small}, each  loop $P_i$ is defined by a sequence of corners with as a 2-dim coordinates:

\begin{small}
\begin{equation}
    P_i=\{C_{i,1},C_{i,2},...,C_{i,N_i}|C_{i,j}\in R^2\}.
\end{equation}
\end{small}

\noindent\textbf{Latent Encoder.}
Given a floorplan sample \begin{small}$P^G_t$\end{small} and corresponding Layout-Init \begin{small}$P^I_t$\end{small} at time $t$, every room/door corner \begin{small}$C_{i,j}^t$\end{small} is mapped to \begin{small}$\hat{C}^t_{i,j}$\end{small} through the Layout-Init and the Noisy encoder:
\begin{small}
  \begin{equation}
  \label{equ:linear_forward}
      \mbox{Linear}([\mbox{AU}(C^t_{i,j})_{G},\mbox{AU}(C^{t}_{i,j})_{I},R_i,\mathbf{1}(i),\mathbf{1}(j),t]).
  \end{equation}
\end{small}
Following the method in \cite{shabani2023housediffusion}, AU enhances relational inference by inserting angular coordinates.
$R_i$ represents a 25-dimensional one-hot vector indicating room type. $1(\cdot)$ refers to a 32-dimensional one-hot vector for the room index $i$ and a corner index $j$. The scalar $t$ is included, and a linear layer transforms the embedding into a 512-dimensional vector.

\noindent\textbf{Continuous Decoder.}
Our reverse process architecture incorporates a sequence of attention layers with structured masking to process embedding vectors \begin{small}$\hat{C}^t_{i,j}$\end{small}.
Each embedding vector \begin{small}$\hat{C}^t_{i,j}$\end{small} undergoes attention layers, which consist of three distinct types of attention: Component-wise Self Attention (CSA), Global Self Attention (GSA) and Relational Cross Attention (RCA).
Following the attention layers, a final linear layer is used to infer noise \begin{small}$\epsilon (C_{i,j},t)$\end{small}, aiding in the iterative refinement of coordinates.

\noindent\textbf{Discrete Decoder.}
Our approach aims to capture geometric relationships (such as collinearity, orthogonality, or corner adjacency), which are difficult to maintain through continuous coordinate regression alone.
As in \cite{shabani2023housediffusion}, we infer coordinates in discrete form. 
After obtaining \begin{small}$C^{t-1}_{i,j}$\end{small} and \begin{small}$C^0_{i,j}$\end{small} from the denoising process, we map \begin{small}$C^0_{i,j}$\end{small} back to the range [0,255] through an affine transformation, round the values, and convert them to binary representation using an “int2bit” function, resulting in an 8-dimensional binary vector.
For each corner, we apply a linear transformation to generate a 512-dimensional embedding vector:
\begin{small}
  \begin{equation}
  \label{equ:discrete}
  \mbox{Linear}([\mbox{AU}(C^0_{i,j}),\mbox{AU}(C^{0}_{i,j})_I,\mbox{int2bit}(C^{0}_{i,j}),R_i,\mathbf{1}(i),\mathbf{1}(j),t]).
    \end{equation}
\end{small}

To enhance network performance, we pass \begin{small}$C^0_{i,j}$\end{small} as the input and use the binary representation from the int2bit function along with conditions, thus providing an initial discrete representation to guide learning. We then apply two additional blocks of attention with structured masking, followed by a final linear layer to produce the 8-dimensional \begin{small}$C^0_{i,j}$\end{small} vector. During testing, we apply binary thresholding to \begin{small}$C^0_{i,j}$\end{small} to retrieve integer coordinates.

\begin{table*}[ht]
\caption{Generation performance of different methods.}
\vskip -0.1in
  \label{table:all_1}
  \begin{center}
    \begin{tabular}{c |c |c c c c| c c c c}
      \toprule
       Method & Realism~$\uparrow$ & \multicolumn{4}{c|}
      {Diversity~$\downarrow$} & \multicolumn{4}{c}{Compatibility~$\downarrow$}\\
      \midrule
       Task & 6 & 5 & 6 & 7 & 8 & 5 & 6 & 7 & 8 \\
      \midrule
      House-GAN\cite{nauata2020house} & -0.95 & 37.5 & 41.0 & 32.9 & 66.4 & 2.5 & 2.4 & 3.2 & 5.3 \\
      House-GAN++\cite{nauata2021house} & -0.52 & 30.4 & 37.6 & 27.3 & 32.9 & 1.9 & 2.2 & 2.4 & 3.9 \\
      HouseDiffusion\cite{shabani2023housediffusion} & -0.19 & 11.2 & 10.3 & 10.4 & 9.5 & 1.5 & 1.2 & 1.7 & 2.5 \\
       \midrule
      PuzzleFusion\cite{hossieni2024puzzlefusion} & - & \multicolumn{4}{c|}{10.55} & \multicolumn{4}{c}{0.97} \\
       \midrule
       Tell2Design~\cite{leng2023Tell2Design} & -1.03 & \multicolumn{4}{c|}{42.74}&\multicolumn{4}{c}{-} \\
       \midrule
      \textbf{Ours} & \textbf{-0.03} & \textbf{8.6} & \textbf{7.5} & \textbf{8.1} & \textbf{9.0} & \textbf{0.24} & \textbf{0.25} & \textbf{0.28} & \textbf{0.32}\\
      \bottomrule
    \end{tabular}
  \end{center}
\end{table*}

\subsubsection{Training Loss}
\label{sec:Training loss}
We train our model with the simple L2-norm regression loss by~\cite{ho2020denoising}.
The loss function consists of noise prediction loss and corner coordinate regression loss.
Specifically, \begin{small}$\epsilon_\theta (C_{i,j},t)$\end{small} is compared with the ground-truth noise to quantify the discrepancy between them.
The discrepancy is typically measured using the masked mean squared error as follows:
\begin{small}
\begin{equation}
    Loss_{n} = \dfrac{\sum_{i=1}^N \Vert \epsilon_\theta(C_{i,j}, t) - \epsilon \Vert_2}{N},
\end{equation}
\end{small}
where $N$ represents total number of samples.
\begin{small}$C_{i,j}$\end{small} is compared with the ground-truth corner coordinate in binary representation:
\begin{small}
\begin{equation}
    Loss_{r} = \dfrac{\sum_{i=1}^N \Vert C^s_{i,j}-C^t_{i,j}\Vert_2}{N}.
\end{equation}
\end{small}
The inference of \begin{small}$C_{i,j}$\end{small} becomes accurate only near the end of the denoising process. Therefore, the discrete branch is used during training only when \begin{small}$C_{i,j}$\end{small}.

\section{Experiments}
\label{sec:experiment}

\subsection{Datasets and implementation details}
\label{sec:Datasets and implementation details}
Our method is implemented in PyTorch and runs on an NVIDIA Quadro RTX 8000. The batch size is set to 128 and the learning rate is 1e-3. Our models are optimized by Adam~\cite{kingma2014adam} with weight decay~\cite{loshchilov2017decoupled} for 250K steps, with the initial learning rate is 1e-3. The network architecture can be found in the Appendix.

\noindent\textbf{Datasets.}
We use the public dataset RPlan~\cite{wu2019data} for experiments. RPlan represents the largest dataset for floorplan with over 80K images. We divide RPlan into five groups (5, 6, 7, 8 rooms) according to the number of rooms for cross-validation.
Ablation experiments were conducted on the 6-room task.
To create paired groups for training, we extracted 10K initial solutions from the LLM using the prompting method from Section~\ref{sec:initialsolutiofromllm}. GPT-4o was used as the LLM in our approach.

\noindent\textbf{Metrics.}
Following prior work~\cite{nauata2020house}, we employ Realism, FID, and Compatibility as evaluation metrics. Realism is an estimate based on a user survey, considering the user's subjective experience. Diversity is measured by the FID score, defined as the Fréchet Inception Distance\cite{heusel2017gans} between the two Gaussian distributions. It is used to measure the similarity between the generated images and the real image. The Compatibility score\cite{abu2015exact} is based on a graph editing distance between the layout sample and the Layout-Final. It is used to measure the distance between the generated layout and the ground truth.
In addition, Macro IoU and Micro IoU are introduced for comparison with the Text-to-Layout method Tell2Design~\cite{leng2023Tell2Design}.
Macro IoU computes the average IoU across different room types, whereas Micro IoU calculates the global IoU by aggregating the IoUs of all rooms.

\subsection{Experimental results}
\label{sec:exper_RPlan}

\begin{figure*}
  \centering
     \includegraphics[width=\linewidth]{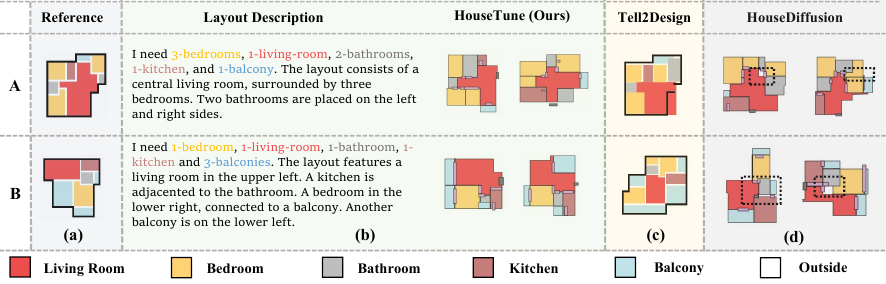}
     \caption{Generation samples from Tell2Design, HouseDiffusion and \ours. The results of \ours\ align well with user requirements in terms of room count and type, with plausible room arrangement. The results produced by Tell2Design exhibit high similarity to the Reference layouts, demonstrating limited diversity. However, the number of generated houses deviates from the Reference, indicating inconsistencies in quantity. HouseDiffusion performs reasonably well; yet, in some cases, it generates gaps or holes within the house (indicated by the dashed-rectangles), along with misaligned object placement, which makes the layout unrealistic.}
     \label{fig:result1}
  \hfill
\end{figure*}

Table~\ref{table:all_1} presents the main results, where we reproduce the reported performance of existing methods for fair comparison. As shown in the table, our approach consistently outperforms prior methods across all evaluation metrics. Specifically, compared to HouseDiffusion, our method improves diversity by 28\% and compatibility by 79\%, demonstrating its effectiveness in generating flexible and spatially efficient layouts.
Tell2Design is evaluated using the IoU metric, with 54.34\% Micro IoU and 53.30\% Macro IoU.
In comparison, \ours, when using unlabeled data, achieves a Micro IoU of 21.84\% and a Macro IoU of 17.75\%, values that are significantly lower than the results of Tell2Design with labeled data.
However, due to fundamental differences in task objectives, a direct comparison based solely on IoU would be unfair.
Tell2Design employs an autoregressive Seq2Seq framework, in which generation is strictly dependent on sequence order.
This results in high IoU scores but significantly limits diversity (see Fig.\ref{fig:result1}(B)), as it demands a large amount of labeled data to capture layout variations. In contrast, our non-autoregressive approach inherently mitigates these limitations, leading to a 79.88\% improvement in diversity compared to Tell2Design.
Furthermore, Tell2Design relies on a costly training process involving artificial pre-training followed by fine-tuning on human annotations. In contrast, our LLM-driven method eliminates the need for manual annotation while still achieving a higher user absolute preference score~(-0.03 v.s. -1.03).
Notably, under the ``Training on Artificial Instructions Only" setting—comparable to our setup—\ours\ surpasses Tell2Design, with a Micro IoU of 15.69, compared to 9.13 for Tell2Design (+71.85\%), and a Macro IoU of 11.43, compared to 6.06 (+88.61\%).
For realism evaluation, we follow the same procedure as House-GAN++ to ensure comparability. User surveys indicate that 65\% of participants perceive our generated layouts as comparable to the ground truth. Implementation details are in the Appendix.

Figure~\ref{fig:result1} presents a visual comparison of layouts generated by \ours, Tell2Design, and HouseDiffusion. The layout descriptions for \ours, Tell2Design’s annotations, and HouseDiffusion’s graphs are all derived from the reference samples shown in Fig.\ref{fig:result1}(a).
Due to the limitation of the plain Seq2Seq model, Tell2Design often produces inaccurate room counts, resulting in unrealistic layouts (see Fig.\ref{fig:result1}(B)).
In contrast, \ours\ and HouseDiffusion explicitly enforce these constraints, producing more structured and diverse layouts.
HouseDiffusion, despite generally producing reasonable layouts, often exhibits misaligned and misplaced objects.
This issue stems from its conditioning strategy, which applies constraints only during the denoising phase, making it less effective in maintaining spatial consistency throughout the generation process (see our ablation study for further analysis).
By contrast, \ours\ refines an initial LLM-generated layout using a diffusion model with conditioning applied in both the noise addition and denoising stages. This dual-phase conditioning enhances structural coherence and spatial accuracy, leading to more practical and well-organized house layouts.

\subsection{Ablation study}
\label{sec:Ablation study}

\begin{figure}
  \centering
     \includegraphics[width=0.5\linewidth]{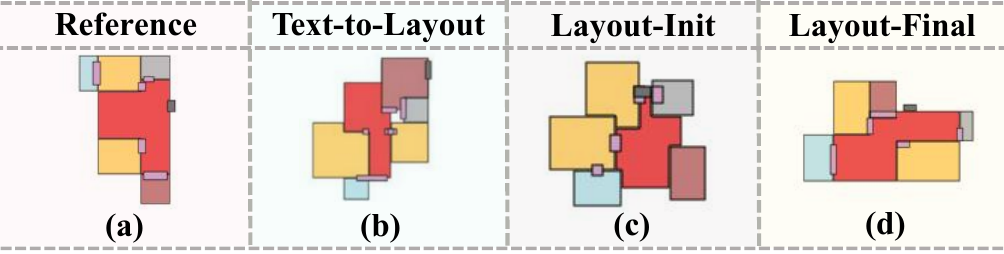}
     \caption{Effect of two stages in house layouts generation. (a) reference; (b) the one-stage method that maps text directly to layout; (c) LLM only, and (d) our method.}
     \label{fig:result2}
\end{figure}

\noindent\textbf{One-stage v.s. two-stage Methods.}
We compare the one-stage Text-to-Layout approach using the text from LLM with our two-stage method. Figure~\ref{fig:result2} presents the comparison results under identical generation conditions. As shown in the figure, the one-stage method frequently produces overlapping rooms and fails to maintain a reasonable spatial distribution.
This limitation arises because textual descriptions alone do not provide sufficient geometric constraints that enable the LLM to generate well-structured layouts.
Additionally, we conduct a quantitative evaluation of both the one-stage and two-stage methods.
Table~\ref{table:ablation_1} presents the main quantitative evaluation results.
Diversity indicates that the distribution of Text-to-Layout method result differs significantly from RPlan.
This is due to a mismatch between the text dataset complexity and task, requiring extensive data support. This limitation restricts the model's generalization ability, preventing it from effectively learning and capturing the true distribution of RPlan. The details of Text-to-Layout are in the Appendix.

\begin{table}[t]
\caption{Comparison between one-stage and two-stage methods. }
\vspace{-0.1in}
  \label{table:ablation_1}
  \begin{center}
    \begin{tabular}{c|c c}
      \toprule
       Method & Diversity~$\downarrow$ & Compatibility~$\downarrow$\\
       \midrule
       1-stage: Text-to-Layout & 109.5 & 10.6\\
       2-stage: \ours\ & \textbf{7.5} & \textbf{0.25}\\
      \bottomrule
    \end{tabular}
  \end{center}
\end{table}

When comparing the Layout-Init and Layout-Final, we can observe that former produces many overlapping rooms and fails to maintain a reasonable distribution; See Figure~\ref{fig:result2}.
These limitations suggest that while the LLM-generated initial layout can capture the general structure of a house layout, it cannot enforce fine-grained spatial constraints. This is where the need for a refinement step becomes essential.

\begin{table}[t]
    \centering
    \caption{Conditional diffusion ablation study.}
    \label{table:ablation_3}
    \begin{subtable}{0.45\textwidth}
        \centering
        \caption{Effect of forward v.s. reverse conditioning.}
        \label{table:ablation_3a}
        \begin{tabular}{cc|cc}
      \toprule
         Forward &  Reverse & Diversity$\downarrow$ & Compatibility$\downarrow$\\
       \midrule
       $\checkmark$&   & 8.57 & 24.79e-2\\
         &  $\checkmark$  & 14.17 & 24.65e-2\\
      $\checkmark$ & $\checkmark$ & 7.47 & 25.44e-2\\
            \bottomrule
        \end{tabular}
    \end{subtable}
    \vspace{4pt}
    \begin{subtable}{0.45\textwidth}
        \centering
        \caption{Effect of different conditions on the generated results.}
        \label{table:ablation_3b}
        \begin{tabular}{cccc}
        \toprule
    \multicolumn{2}{c|}{Rate} &  Macro IoU~$\uparrow$ & Micro IoU~$\uparrow$\\
      \midrule
       \multicolumn{2}{c|}{1e-1} & 21.37\% &18.46\%\\
       \multicolumn{2}{c|}{1e-2} & 20.12\% & 16.59\%\\
       \multicolumn{2}{c|}{1e-3} & 18.67\% & 15.43\% \\
            \bottomrule
        \end{tabular}
    \end{subtable}
\end{table}

\begin{figure}
  \centering
     \includegraphics[width=0.5\linewidth]{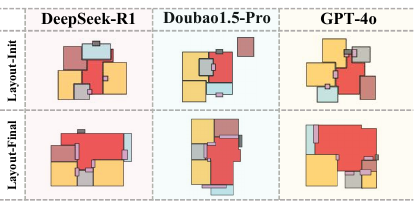}
     \caption{Generation results with different LLMs.}
     \label{fig:result3}
\end{figure}

\noindent\textbf{Robustness and effectiveness of the prompts.}
Our method maintains consistent quality across both commercial and open-source LLMs, namely DeepSeek-R1 and Doubao-1.5Pro. As shown in Fig.\ref{fig:result3} and Table.\ref{table:ablation_2a}, the generated JSON representations preserve structural integrity across different models, enabling seamless downstream diffusion refinement without additional fine-tuning. This demonstrates that our prompting strategy exhibits low dependency on proprietary models.
Furthermore, Table~\ref{table:ablation_2b} presents the ablation results for different prompt designs. \textbf{P1} employs only the task definition prompt to generate semantic representations. \textbf{P2} incorporates a common-sense prompt, improving reasoning accuracy. \textbf{P3} adopts a basic Chain-of-Thought (CoT) prompt (“Let’s think step by step”~\cite{kojima2022large}), further enhancing performance. \textbf{P4} introduces explicit reasoning chains, leading to additional improvements over P3. These results indicate that our tailored prompting strategy effectively enhances the quality of the initial layout generation.

\noindent\textbf{Conditional diffusion ablation study.}
Unlike the diffusion model used in HouseDiffusion, our approach integrates conditional constraints in both the forward and reverse processes. To evaluate the effectiveness of this design, we analyzed the impact of each constraint by independently varying their proportions. As shown in Table~\ref{table:ablation_3a}, incorporating conditional information during the noise addition phase leads to better performance compared to applying it only during the denoising phase. This is because early integration of conditions allows the model to establish condition-related features at an earlier stage.
A single-stage conditioning approach may result in weaker constraints, leading to errors in room sizes and spatial relationships. The best performance is achieved when conditions are applied in both phases, confirming the effectiveness of the dual-phase conditioning strategy.

Furthermore, we examined the effect of different conditional participation ratios on generation quality, as presented in Table~\ref{table:ablation_3b}. The results indicate that as the conditional ratio increases, model performance gradually declines, suggesting that an excessive amount of conditional information causes the model to diverge from the true data distribution. Therefore, properly regulating the conditional ratio is crucial for maintaining the accuracy and realism of the generated samples.

\begin{table}[t]
    \centering
    \caption{Prompt ablation study.}
    \label{table: ablation_2}
    \begin{subtable}{0.4\textwidth}
        \centering
        \caption{Comparative analysis of different LLMs.}
        \label{table:ablation_2a}
    \begin{tabular}{c| c c }
    \toprule
        LLMs  & Diversity~$\downarrow$ &  IoU~$\uparrow$\\
    \midrule
      DeepSeek-R1 & 94.3& 22.3\% \\
      DouBao-1.5Pro & 95.6& 20.2\% \\
    \midrule
       GPT-4o & 95.1& 21.6\% \\
    \bottomrule
    \end{tabular}
    \end{subtable}
    \vspace{4pt}
    \begin{subtable}{0.47\textwidth}
        \centering
        \caption{Comparative analysis of different prompting strategies.}
        \label{table:ablation_2b}
        \begin{tabular}{ c | c c c }
     \toprule
     &Description & Diversity~$\downarrow$ & IoU~$\uparrow$\\
    \midrule
    P1&Task definition prompt & 98.6 & 12.36\%\\
    P2&Common sense prompt & 92.3 & 14.57\% \\
    P3&Think step by step & 86.5 & 20.69\% \\
    P4&Predefined CoT prompt & 75.2 & 21.41\% \\
    \bottomrule
    \end{tabular}
    \end{subtable}
\end{table}

\section{Conclusion}
\label{sec:conclusion}
This paper proposes a two-stage text-to-floorplan generation model, enabling user-friendly house layout generation. This work guides the reasoning of the LLM through Chain-of-Thought prompting to generate an initial layout based on user requirements. The initial layout is then refined using a diffusion model to produce the final house layout. Experimental results show that our method achieves state-of-the-art performance. Given the potential of LLMs, we expect that our solution be extended to the generation of other complex architectural designs such as shopping malls and office buildings. This will be a subject of our future study.

\bibliographystyle{unsrtnat}
\bibliography{references} 

\end{document}